 \documentclass[pmlr,twocolumn]{jmlr} % W&CP article

\usepackage{booktabs}
\usepackage[font=small,skip=2pt]{caption}

\usepackage[load-configurations=version-1]{siunitx} % newer version
\theorembodyfont{\upshape}
\theoremheaderfont{\scshape}
\theorempostheader{:}
\theoremsep{\newline}

 \jmlrvolume{ML4H Extended Abstract Arxiv Index}
\jmlryear{2020}
\jmlrsubmitted{2020}
\jmlrpublished{}
\jmlrworkshop{Machine Learning for Health (ML4H) 2020}

\title[Effectiveness of Batch-Norm Statistics to mitigate catastrophic forgetting]{The unreasonable effectiveness of Batch-Norm statistics in addressing catastrophic forgetting across medical institutions}

\author{%
\begin{center}
\Name{Sharut Gupta$^1$}, \Name{Praveer Singh$^1$}, \Name{Ken Chang$^1$}, \Name{Mehak Aggarwal$^1$}, \Name{Nishanth Arun$^1$}, \Name{Liangqiong Qu$^3$}, \Name{Katharina Hoebel$^{1,2}$}, \Name{Jay Patel$^{1,2}$}, \Name{Mishka Gidwani $^1$}, \Name{Ashwin Vaswani$^1$},
\Name{Daniel L Rubin$^{3}$},
\Name{Jayashree Kalpathy-Cramer$^1$} 
\\
\addr $^1$ Athinoula A. Martinos Center for Biomedical Imaging, Boston, MA, USA
 \\ $^2$ Massachusetts Institute of Technology, Cambridge, MA, USA
 \\ $^3$ Department of Radiology and Biomedical Data Science, Stanford University, Palo Alto, CA, USA
\end{center}
}
\begin{document}
\maketitle
\begin{abstract}
Model brittleness is a primary concern when deploying deep learning models in medical settings owing to inter-institution variations, like patient demographics and intra-institution variation, such as multiple scanner types. While simply training on the combined datasets is fraught with data privacy limitations, fine-tuning the model on subsequent institutions after training it on the original institution results in a decrease in performance on the original dataset, a phenomenon called catastrophic forgetting. In this paper, we investigate trade-off between model refinement and retention of previously learned knowledge and subsequently address catastrophic forgetting for the assessment of mammographic breast density. More specifically, we propose a simple yet effective approach, adapting Elastic weight consolidation (EWC) using the global batch normalization (BN) statistics of the original dataset. The results of this study provide guidance for the deployment of clinical deep learning models where continuous learning is needed for domain expansion.   
\end{abstract}

\section{Introduction}
\label{sec:intro}
Mammographic breast density assessment is routinely used to predict breast cancer risk, decreasing the risk of breast cancer mortality \cite{tabar2001beyond,razzaghi2012mammographic}. 
% The current criteria for mammographic breast density classification is based on the Breast Imaging Reporting and Data System (BI-RADS), which divides breast density assessment into four density categories: fatty, scattered, heterogeneously dense, extremely dense \cite{liberman1998breast}. In practice however, BI-RADS is highly subjective, resulting in high inter-rater variability which may confer undue patient anxiety and unnecessary supplemental screening examinations \cite{sprague2016variation}. This human variation recommends the addition of computer-based methods. \\
Deep learning models have shown state of the art performance for many medical imaging tasks \cite{esteva2017dermatologist,chang2019automatic,li2020siamese}. Previous works that developed deep learning algorithms for breast density assessment have only focused on a single institution with a single digital mammography system \cite{lehman2019mammographic,mohamed2018deep}. A major hurdle to large-scale clinical deployment for a deep learning based breast density assessment tool is the poor generalizability across different institutions and scanner types owing to variability in patient demographics, disease prevalence, and imaging acquisition techniques \cite{zech2018variable}.
Traditionally, deep learning models have been extended to new medical datasets after training on a related dataset. However, simply fine-tuning models on new institutions is insufficient, as this can induce catastrophic forgetting (CF). This phenomenon occurs when models do not preserve previously learned knowledge, degrading performance on the original dataset \cite{goodfellow2013empirical}. This poses a major challenge for regulatory agencies such as the Food and Drug Administration (FDA) because deep learning models that have been fine-tuned since their approval may no longer have threshold performance on the original test set. In this work, we explore techniques to mitigate CF when fine-tuning on new medical datasets, otherwise known as domain expansion. For simplicity, the original domain is referred to as Dataset O while the target domain is referred as Dataset T. The key contributions of this study are as follows: 1) We propose a very simple yet effective technique to avoid CF, by utilizing the global BN statistics of Dataset O~\footnote[0*]{*In this paper, global BN statistics of a dataset will refer to the running mean and standard deviations of BN layers computed using that dataset} instead of Dataset T when fine-tuning on T.  2) We show the efficacy of this technique under two different scenario's: first restricting fine-tuning to only BN layers (motivated by \cite{karani2018lifelong}) and second fine-tuning using all the layers 3) We lastly show how a very popular continuous learning algorithm (EWC \cite{kirkpatrick2017overcoming}) fails miserably when trained on large datasets using complex architectures, and then highlight how adapting EWC with our technique helps not only in successful domain expansion over T but also mitigates CF on O.

\vspace{-2mm}
\section{Method}
\subsection{Datasets}
Digital screening mammograms from a multi institutional dataset with 5 digital mammography systems (Dataset O) and from another institution (Dataset T) are used. All images were interpreted by a single radiologist using ACR BI-RADS breast density lexicon (Category A: fatty, Category B: scattered, Category C: heterogeneously dense, Category D: extremely dense)\cite{liberman1998breast}. 
%Dataset A is further split into Dataset C and Dataset D which differ in the mode of acquisition.
\begin{figure}[htbp]
 % Caption and label go in the first argument and the figure contents
 % go in the second argument
\floatconts
  {fig:bn_fig}
  {\caption{The two experimental BN approaches for fine-tuning on target dataset (T): (Top) Using global statistics of T; (Bottom) Using global statistics of the original dataset (O). For both approaches, we tested by fine-tuning with both: only the BN layers as well as all the layers.}}
  {\includegraphics[width=1.0\linewidth]{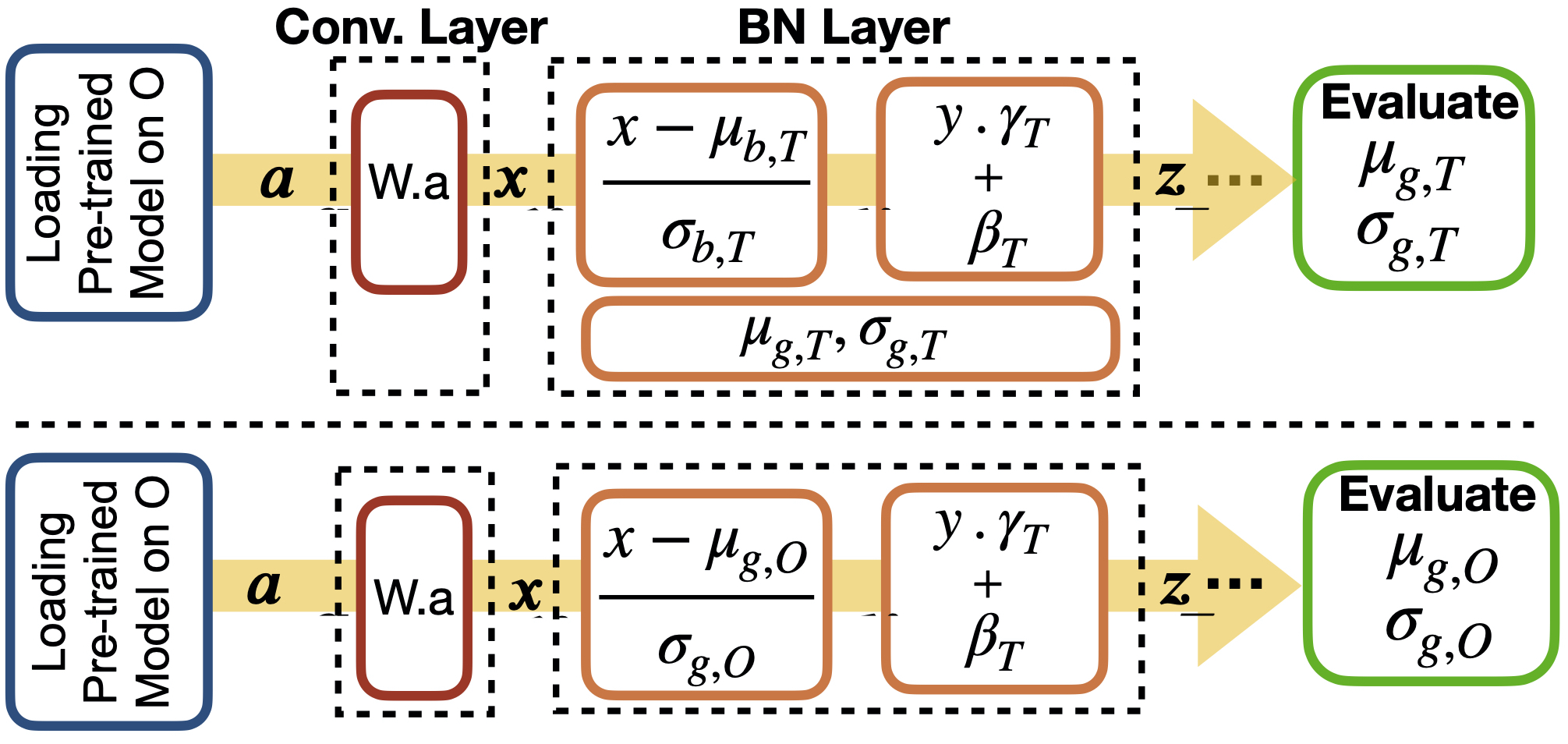}}
% \caption{This is a figure.}
\end{figure}

\subsection{Fine-tuning only BN Layers with and without EWC}
BN has become a popular technique to regulate shifts in the distribution of network activations during training  \cite{ioffe2015batch}. Fig ~\ref{fig:bn_fig} (top row) illustrates the pipeline for traditional fine-tuning on Dataset T wherein BN layer normalises the input using Dataset T's batch statistics  ($\mu_{b,T}$ and $\sigma_{b,T}$) during training, and performs inference with its global statistics ($\mu_{g,T}$ and $\sigma_{g, T}$) which were calculated while training. 
With the intuition of aligning the data distribution of datasets T and O, and retaining the original model performance, we test run several experiments by fine-tuning only BN layers on T while freezing all other layers. In addition to this standard approach of BN, we fine-tune and evaluate using the global BN statistics of Dataset O (as shown in bottom row of Fig \ref{fig:bn_fig}). The BN (both for fine-tuning and inference) on an incoming batch in such a case is represented by: $y_{i} = \gamma\frac{x_i-\mu_{g,O}}{\sqrt{{\sigma^{2}_{g,O}}+\epsilon}} +\beta$

% \begin{equation} \label{eq:BNO}
%  y_{i} = \gamma\frac{x_i-\mu_{g,O}}{\sqrt{{\sigma^{2}_{g,O}}+\epsilon}} +\beta
% \end{equation}

Furthermore, we employ EWC in the training pipeline while fine-tuning the BN layers on Dataset T with global BN statistics of O.

\subsection{Fine-tuning all layers with and without EWC}
With the primary objective of achieving best performance on Dataset T while retaining the original performance on Dataset O, we further investigate the effects of fine-tuning all layers (BN as well as convolutional) with and without EWC. Here too, we experiment with both, the traditional BN technique using the global statistics of T (Fig \ref{fig:bn_fig} top) v/s our technique of using global statistics of O (Fig\ref{fig:bn_fig} bottom). Utilizing EWC, effectively constrains the updates of those weights which are most important for Dataset O, thus ultimately preventing catastrophic forgetting for O. The effective loss function $L(\theta)$ for elastic weight consolidation is given by 
% \begin{equation}
%     L(\theta) = L_{Dataset T}(\theta) + \sum_i \frac{\lambda}{2}F_{i}(\theta_{i}-\theta^{*}_{Dataset O,i})^{2}
% \end{equation}
$L(\theta) = L_{Dataset T}(\theta) + \sum_i \frac{\lambda}{2}F_{i}(\theta_{i}-\theta^{*}_{Dataset O,i})^{2}$ 
where $L_{Dataset T}(\theta)$ is the loss function for Dataset T, $\theta^{*}_{Dataset  O}$ represents the optimal model parameters for Dataset O and i iterates over all the model parameters. The parameter $\lambda$ is used as a trade off between the relative importance of Datasets O and T \cite{kirkpatrick2017overcoming}.
% As a result, a constraint on some (not all) model parameters, allows the solution to stay in a low-error region which is optimal for both the datasets \cite{kirkpatrick2017overcoming}.

\section{Results}
\subsection{Baseline Experiments}
We train the baseline deep learning model using three different constructs: 1) training solely on Dataset O, 2) training on Dataset O and fine-tuning on Dataset T, and 3) training on the combined datasets O and T.

\begin{figure}[htbp]
 % Caption and label go in the first argument and the figure contents
 % go in the second argument
\floatconts
  {fig:baseline}
  {\caption{Performance results for the baselines }}
  {\includegraphics[width=1\linewidth]{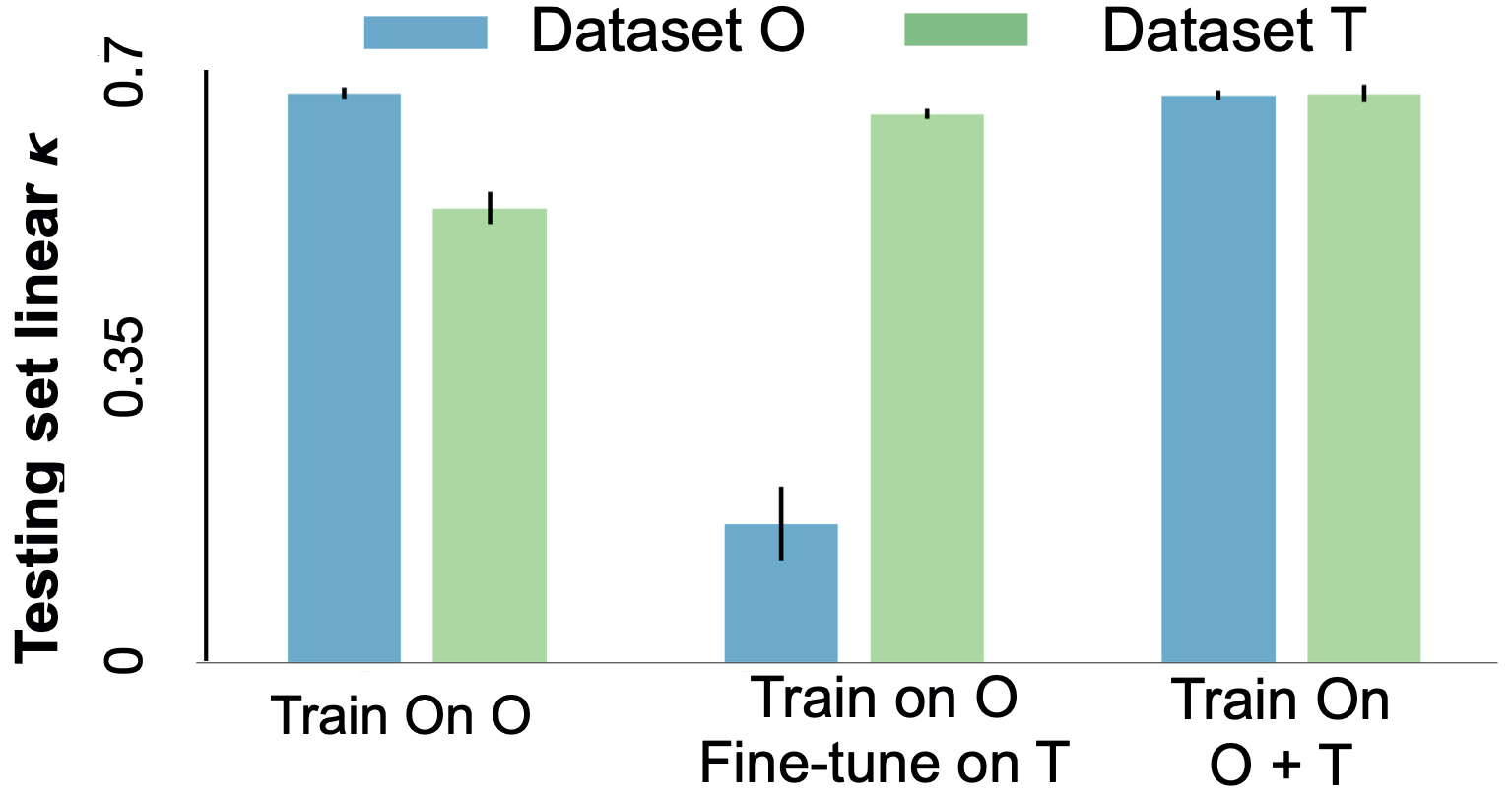}}
\end{figure}

As shown in Fig \ref{fig:baseline}, models exclusively trained on Dataset O ($\kappa$: 0.6694) do not generalise well on Dataset T ($\kappa$: 0.5621). When the model is trained on Dataset O and fine-tuned on Dataset T, it abruptly forgets the information it learnt on Dataset O ($\kappa$: 0.1635, p $<$ 0.01). Only when the model is trained on both datasets O and T simultaneously does it achieve high performance on both domains($\kappa$: 0.6671, 0.6690 on O and T respectively).
%using Dataset A as original dataset O and Dataset B as target dataset T, or using Dataset C as O and Dataset D as T,  

\subsection{Fine-tuning only BN Layers with and without EWC}
All except BN layers are frozen to fine-tune the model on Dataset T and  performance on both datasets is evaluated after fine-tuning on Dataset T with varying importance parameters  ($\lambda$)  as  depicted  in  Fig \ref{fig:lineplot}(left). We observe that at $\lambda=0$ (conventional fine-tuning without EWC), when the global BN statistics of Dataset T are used, the model undergoes a large reduction (p $<$ 0.01) in the performance on Dataset O. Moreover, the performance on Dataset T is lower (p $<$ 0.01) compared to the performance when fine-tuned with all the layers. In contrast, when the model is fine-tuned using the global BN statistics of Dataset O, we see a recovery (p $<$ 0.01) in the performance on Dataset O as well as an increase in the performance on Dataset T from the baseline model before fine-tuning (p $<$ 0.01). This is further substantiated by plotting UMAPS for Dataset O (Fig \ref{fig:umaps} (a),(b) and (c) in appendix), wherein feature points of the model trained with global statistics of T seem intermingled, while for those trained with global statistics of O has feature points well aligned along a particular direction. (similar to original model trained purely on Dataset O). %Similar results are obtained for Dataset C (O) and Dataset D(T).

Clearly, fine-tuning BN layers using the global BN statistics of Dataset O confers a performance advantage over fine-tuning using the global BN statstics of Dataset T. We also observe that when using the global statistics of Dataset O, at $\lambda=0.05$, we achieve maximum performance on both Dataset O ($\kappa$: 0.6734) and Dataset T ($\kappa$: 0.5940). 
% The performance at $\lambda=0.05$ on Dataset O and Dataset T is $\kappa$: 0.6734 and $\kappa$: 0.5940 respectively.
%on Dataset C (O) is $\kappa$: 0.6291 and on Dataset D (T) is $\kappa$: 0.5944. At the same $\lambda$, the performance 

 \begin{figure}[htbp]
 % Caption and label go in the first argument and the figure contents
 % go in the second argument
\floatconts
  {fig:lineplot}
  {\caption{Performance with varying $\lambda$ when fine-tuning only BN layers (left) and all layers (right)}}
  {\includegraphics[width=1.0\linewidth]{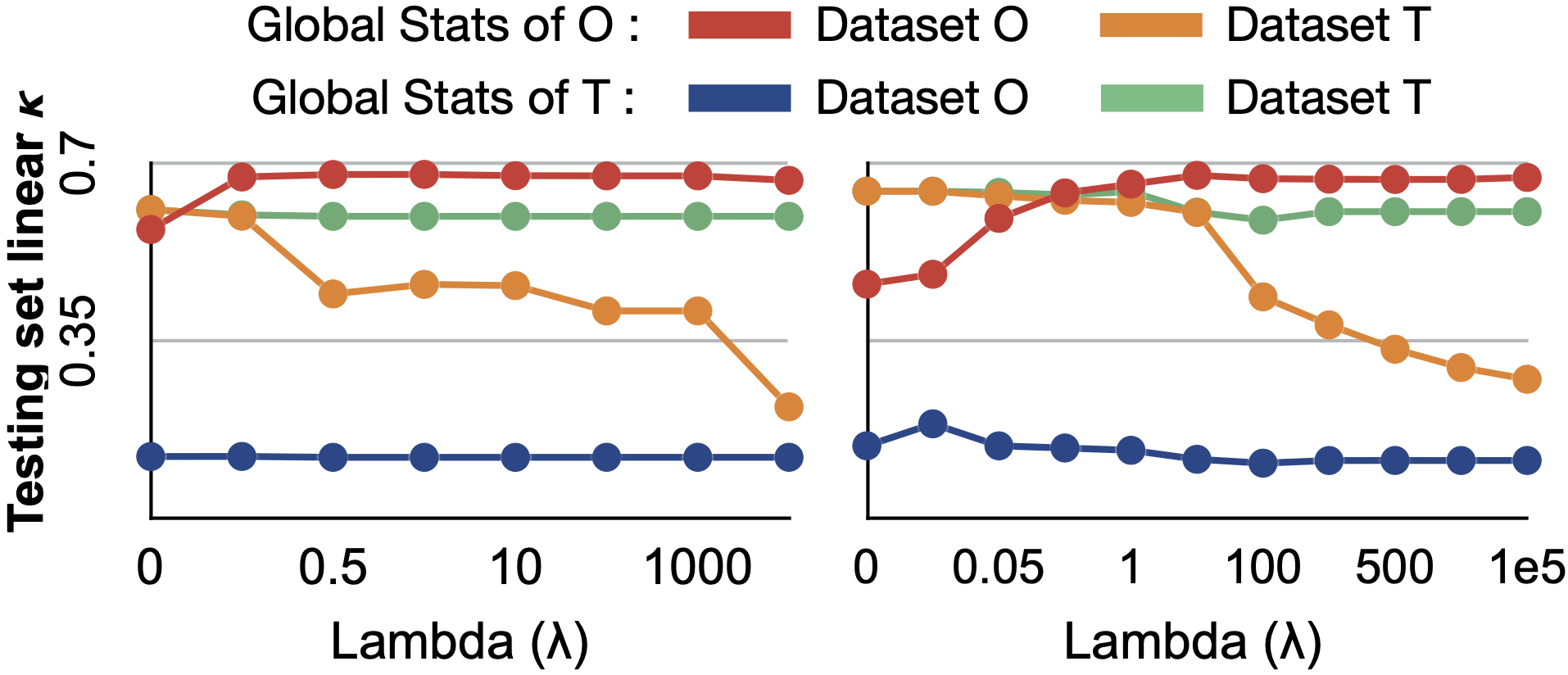}}
  
\end{figure}

\subsection{Fine-tuning all Layers with and without EWC}
Similar to the above setup, model is fine-tuned for all layers on Dataset T and evaluated first using the global BN statistics of Dataset T and second using the global BN statistics of Dataset O. 
% When this model is evaluated using the global BN statistics of Dataset T
In the former, we again observe that performance on Dataset O degrades significantly (p $<$ 0.01), the canonical presentation of CF. Evaluating with global BN statistics of Dataset O attenuates the performance loss (p $<$ 0.01). 
As depicted in fig \ref{fig:lineplot} (Right), while fine-tuning with the global statistics of Dataset T, performance on Dataset O remains consistently sub-optimal with varying lambda. When fine-tuning with the global statistics of Dataset O at $\lambda=1e+5$, catastrophic forgetting is almost entirely mitigated, with a performance of $\kappa$: 0.6676 on Dataset O as compared to the baseline performance of $\kappa$: 0.6694. However, this lambda value prevents the model from learning the features of the target domain, and hence the model performs poorly on Dataset T ($\kappa$: 0.2310). %The best performance on both Dataset C ($\kappa$: 0.6037) and Dataset D ($\kappa$: 0.6662) is obtained at $\lambda=0.005$. 
The best performance on both Dataset O ($\kappa$: 0.6600)and Dataset T ($\kappa$: 0.6215) is achieved at $\lambda=1$. As expected, at the lambda which gives highest performance on both Datasets, fine-tuning BN layers rather than all layers gives better performance on Dataset O (p $<$ 0.01 for both datasets). Conversely, fine-tuning all layers rather than just BN layers results in higher performance on Dataset T (p $<$ 0.01 for both datasets). Fig \ref{fig:umaps} ((d),(e) and (f) in appendix) further illustrate our claims where feature points of model trained with global statistics of T seem intermingled; for those trained with global statistics of O have spatially segregated classes; while the model trained with EWC has feature points well aligned along a particular direction (similar to original model trained purely on Dataset O).

% \vspace{-2mm}
\section{Conclusion}
We demonstrate that using global BN statistics of the original dataset (Dataset O) when fine-tuning on the target dataset (Dataset T), partially alleviates catastrophic forgetting. When used in conjunction with EWC, it completely mitigates catastrophic forgetting. We also observe that fine-tuning BN layers alone is insufficient to achieve high performance on Dataset T. Rather, fine-tuning all layers is necessary to achieve high performance on both Dataset O and T. The efficacy of our approach is illustrated through domain expansion from a large, multi-institutional dataset to a single institution dataset. This has  major implications for the clinical deployment of deep learning models where generalizability remains a major hurdle to large-scale use. In the future we plan to experiment our technique across different imaging acquisition systems and perform the same study in reverse order.

% \acks{This study was supported by National Institutes of Health (NIH) grants U01CA154601, U24CA180927, and U24CA180918 to J. Kalpathy-Cramer, U01CA242879 to D. Rubin and J. Kalpathy-Cramer, and National Science Foundation (NSF) grant NSF1622542 to J. Kalpathy-Cramer. This research was carried out in whole or in part at the Athinoula A. Martinos Center for Biomedical Imaging at the Massachusetts General Hospital, using resources provided by the Center for Functional Neuroimaging Technologies, P41EB015896, a P41 Biotechnology Resource Grant supported by the National Institute of Biomedical Imaging and Bioengineering (NIBIB), National Institutes of Health.}

\bibliography{jmlr-sample}
\appendix
\section{Appendix}\label{apd:first}
\subsection{Dataset Preprocessing}
The Dataset O (n = 103890 and Dataset T (n = 8603) patient cohorts were split into training, validation, and testing sets in a 7:2:1 ratio, on a patient level. The intensity of each image was scaled to be between 0 and 1 by dividing by the maximum value of the image format (4095 for 12 bit and 16383 for 14 bit). To ensure proper input size to the pre-trained neural network architectures, the images were resized to  224x224x3.

\subsection{Training and Prediction}
For the baseline classification model, a Resnet50 architecture with ImageNet pretrained weights \cite{he2016deep} was used. A batch size of 32 images, Cross entropy loss function and Adam optimizer (lr = $10^{-6}$) are used across models. Early stopping with a patience of 20 epochs and real time augmentation is is used to prevent overfitting. All models are evaluated using Cohen's Kappa scores with linear weighting ($\kappa$). A Wilcoxon signed-rank test at a significance level of p = .05 was used for statistical comparisons of model performance.

\begin{figure}[htbp]
 % Caption and label go in the first argument and the figure contents
 % go in the second argument
\floatconts
  {fig:umaps}
  {\caption{UMAPs for Dataset O when (a) Model is trained just on Dataset O; When fine-tuned BN layers with global stats of Dataset (b) T; (c) O; Fine-tuned all layers with global stats of Dataset (d) T; (e) O; (f) O with EWC ($\lambda$=1)}}
  {\includegraphics[width=1\linewidth]{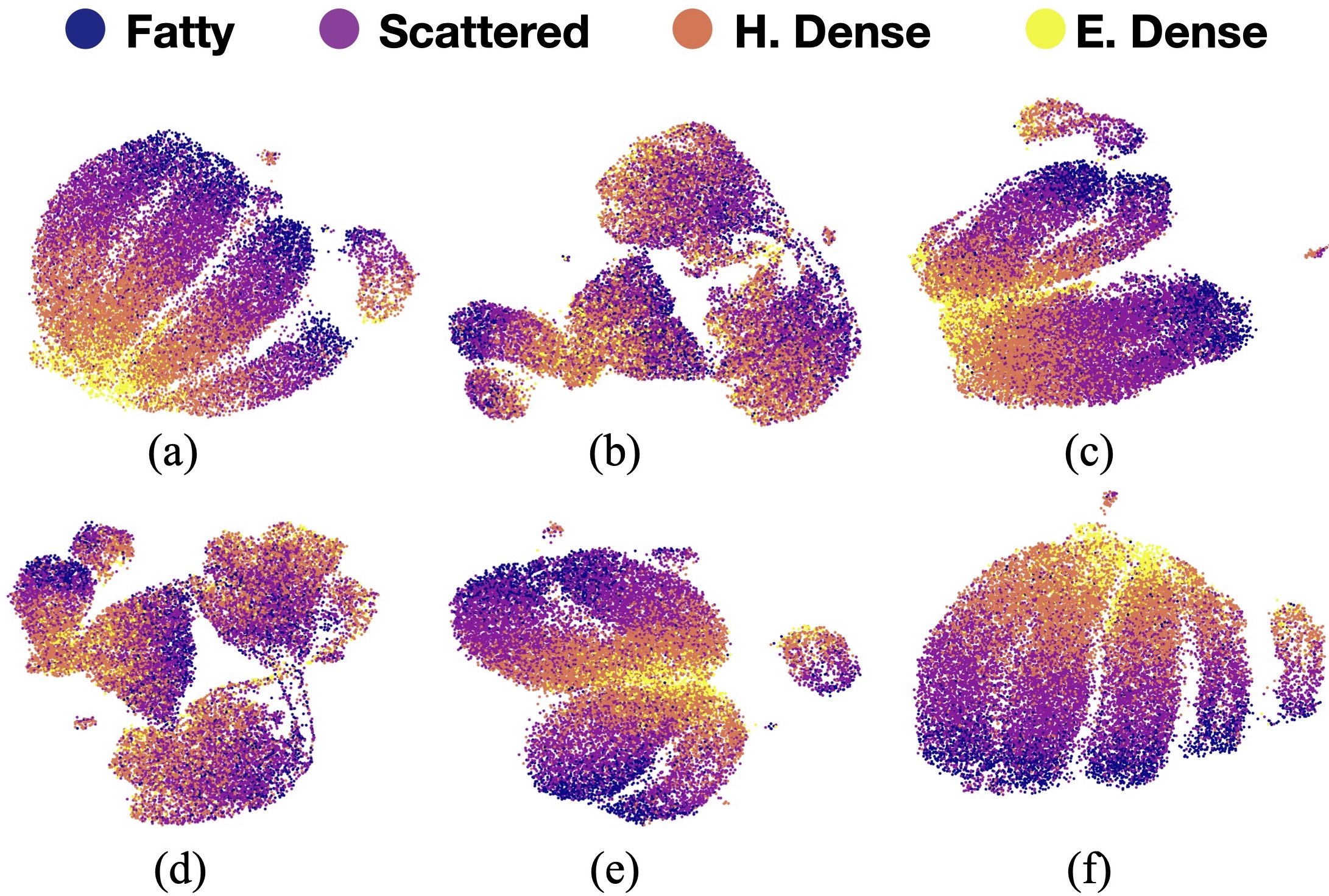}}
\end{figure}

% \begin{figure}[htbp]
%  % Caption and label go in the first argument and the figure contents
%  % go in the second argument
% \floatconts
%   {fig:losscurve}
%   {\caption{Evolution of validation loss during training for Dataset O (left) and Dataset T (right) with varying $\lambda$}}
%   {\includegraphics[width=1\linewidth]{./Figures/loss.jpg}}
% \end{figure}

\end{document}